\documentclass[Journal,letterpaper,NoLineNumbers]{ascelike-new}
\usepackage[utf8]{inputenc}
\usepackage[T1]{fontenc}
\usepackage{lmodern}

\usepackage{graphicx}
\usepackage[style=base,figurename=Fig.,labelfont=bf,labelsep=period]{caption}
\usepackage{float}
\usepackage{subcaption}
\usepackage{amsmath}
\usepackage{newtxtext,newtxmath}
\usepackage[colorlinks=true,citecolor=red,linkcolor=black]{hyperref}
\usepackage{setspace}
\NameTag{Hu, \today}
\begin{document}

\title{Sample-Efficient Robot Skill Learning for Construction Tasks: Benchmarking Hierarchical Reinforcement Learning and Vision-Language-Action VLA Model}

\author[1]{Zhaofeng Hu}
\author[2]{Hongrui Yu}
\author[3]{Vaidhyanathan Chandramouli}
\author[4]{Ci-Jyun Liang*}

\affil[1]{Department of Civil Engineering, Stony Brook University, Stony Brook, NY 11794 USA}
\affil[2]{Department of Civil and Environmental Engineering, Virginia Tech, Blacksburg, VA 24061 USA}
\affil[3]{Department of Electrical and Computer Engineering, Virginia Tech, Blacksburg, VA 24061 USA}
\affil[4]{*Department of Civil Engineering, Stony Brook University, Stony Brook, NY 11794 USA. Email: ci-jyun.liang@stonybrook.edu}

\maketitle

\begin{abstract}
This study evaluates two leading approaches for teaching construction robots new skills to understand their applicability for construction automation: a Vision-Language-Action (VLA) model and Reinforcement Learning (RL) methods. The goal is to understand both task performance and the practical effort needed to deploy each approach on real jobs. The authors developed two teleoperation interfaces to control the robots and collect the demonstrations needed, both of which proved effective for training robots for long-horizon and dexterous tasks. In addition, the authors conduct a three-stage evaluation. First, the authors compare a Multi-Layer Perceptron (MLP) policy with a Deep Q-network (DQN) imitation model to identify the stronger RL baseline, focusing on model performance, generalization, and a pick-up experiment. Second, three different VLA models are trained in two different scenarios and compared with each other. Third, the authors benchmark the selected RL baseline against the VLA model using computational and sample-efficiency measures and then a robot experiment on a multi-stage panel installation task that includes transport and installation. The VLA model demonstrates strong generalization and few-shot capability, achieving 60\% and 100\% success in the pickup phase. In comparison, DQN can be made robust but needs additional noise during tuning, which increases the workload. Overall, the findings indicate that VLA offers practical advantages for changing tasks by reducing programming effort and enabling useful performance with minimal data, while DQN provides a viable baseline when sufficient tuning effort is acceptable. 
\end{abstract}

\section{Practical Applications}
Construction robots can improve jobsite productivity, safety, and delivery timelines. This study compares two leading ways to program these robots: a vision-and-language model that learns from images and motion examples to produce actions, and a Reinforcement Learning approach trained with rewards. Using a cost–benefit lens, the authors examine both task performance and the effort required to set up and tune each method. In robot and computational tests on a two-phase panel task (moving a panel and then installing it), the Vision--Language-Action model showed strong generalization and useful “few-shot” behavior: with minimal training, it reached 60\% and 100\% success in the two different scenes of the pickup process. The Reinforcement Learning approach reached similar pickup success but needed added noise during tuning to achieve robust results, increasing the workload. Each method was evaluated with the input types most suitable for it (vision and action for the vision-and-language model; trajectory and force for the reinforcement-learning model). For practitioners, these findings suggest that Vision--Language-Action systems can reduce programming effort while still achieving high accuracy on installation tasks, whereas reinforcement learning may require more hands-on tuning to reach comparable reliability.

\section{Introduction}
The construction industry is confronting a persistent labor shortfall. In the United States alone, industry analysts estimated that approximately 501,000 additional workers were needed in 2024 just to meet demand, which demonstrates the magnitude of the workforce gap \cite{ABC_2024_workforce_shortage}. This labor scarcity has tangible downstream effects, including empirical work linking negative shocks to construction labor supply with slower residential production and higher new-home prices. Studies also document schedule slippage and cost escalation associated with craft shortages \cite{CII_2018_workforce_development_FR335,Karimi_2018_cost_performance,HowardWangZhang_2024_housing_labor}.
Robotics and automation have long been proposed as pathways to boost productivity and partially substitute for scarce labor, and scholarly reviews in construction settings identify significant potential productivity gains from targeted robotic adoption \cite{Bock_2015_future_construction_automation,Liang_2021_HRC_review}. In recent years, research prototypes and early deployments have spanned a diverse array of on-site tasks: autonomous surveying and layout, drilling, bricklaying, drywall finishing, façade operations, and collaborative manipulation \cite{howard2025industrial,xiao2022recent,armeni2024construction}. The majority of these tasks are manipulation, i.e., using robot arms or manipulators to pick up and place the construction materials \cite{armeni2024construction}, such as bricklaying \cite{mantha2025multi,liuexploiting}, ceiling tile \cite{liang2020teaching}, drywall \cite{wang2021interactive}, and bolting \cite{yasutomi2023dual}.
Yet, long-horizon and complex material manipulation and installation remain comparatively underexplored relative to other tasks, with the literature featuring only scattered case studies and focused prototypes despite the tasks’ multi-stage and complicated nature \cite{gil2013installation,chandramouliconstruction,LiZou2023_AEI_GAIL_Construction}.

Learning-based methods have been recently introduced to construction robots to tackle unstructured, dynamic, and complex manipulation tasks \cite{huang2023imitate,liang2020teaching,yu2024cloud,yu2023mutual}. These methods can relieve the need for extensive robot programming or teleoperating by demonstrating the task to the robot.
Two methodological trends are especially salient for installation-like tasks. First, Reinforcement Learning (RL) and Imitation Learning (IL) are increasingly applied to contact-rich assembly and insertion, where policies must achieve alignment, compliant insertion, and precise placement under uncertainty \cite{Argall2009SurveyLfD,Kober2013RLRoboSurvey}. RL frames robot control as sequential decision-making to maximize expected return, while IL derives policies by training a model to replicate expert demonstrations \cite{Ross2011DAgger}. They can also be combined to have RL-based human skill imitation. Recent studies demonstrate IL- or RL-based peg-in-hole and force-aware assembly with real systems, illustrating the relevance to installation primitives such as align–insert–fasten–seal \cite{Zang2023ProMPsPegInHole,liang2022trajectory,chandramouliconstruction}.

Second, recent advanced robotic foundation models (large, pre-trained Vision-Language-Action (VLA) or generalist manipulation policies) promise general-purpose skills and zero- or few-shot skill imitation. Models such as Robotics Transformer 2 (RT-2) demonstrate zero-shot reasoning from Internet-scale pretraining into robot control; the Open X-Embodiment and RT-X collaboration shows cross-embodiment transfer from multi-robot, multi-task datasets; Octo et al. (2024) provides an open generalist policy that adapts rapidly to new observation/action spaces; and VIMA exhibits strong zero-shot generalization via multimodal prompting \cite{rt22023arxiv,OpenXEmbodiment2023RTX,Octo2024,jiang2022vima}. These traits are attractive for construction tasks, where the demonstration itself can be costly and physically demanding for workers to collect \cite{wang2021interactive}.

Despite the rapid development and advances in robot learning approaches, it is unclear for complex and multi-stage construction tasks, which option would deliver satisfactory performance with minimal demonstration and model training efforts. Some methods require abundant demonstration data to achieve sufficient performance. For instance, the Learning from Demonstration (LfD) method in \citeN{liang2020teaching} required over a thousand demonstration data in the simulator and real-world. There is a lack of systematic benchmarking of RL-based imitation and general-purpose-based one-shot imitation for construction tasks \cite{guruprasad2024benchmarking,duan2016benchmarking}. The RL model would require a lot less data to train compared to establishing a VLA from scratch. Nonetheless, if VLA models can be used for construction tasks with minimal data requirements, it will offer a promising zero-programming pathway for rapid construction robot adoption. 

This paper investigates when and why these advances will be the optimal option for programming robots to perform long-horizon material manipulation and installation with a cost-benefit framework. The authors designed a dual-phase experiment to understand the data requirement for both RL-based and VLA-based imitation models to achieve similar task performance. To elaborate, in this study, the authors (i) develop a learning framework that pairs imitation-initialized policies with reinforcement-based refinement; and (ii) evaluate foundation model initializations for few-shot transfer across materials, geometries, and layouts. Our results aim to clarify which learning-based robotics can help mitigate labor bottlenecks, shorten installation cycle time, and improve quality in long-horizon material workflows on real construction sites.

\section{Literature Review}
This section provides a systematic review of both progress in general construction robot skill learning and innovations in sample-efficient learning models, including Hierarchical Reinforcement Learning (HRL) and VLA.

\subsection{Construction Robot Learning}
Imitation Learning, also known as LfD, addresses the construction skill-transfer problem by learning policies directly from expert behavior in an effective and accessible way. Construction researchers have embraced the algorithmic families, including behavior cloning, interactive IL such as DAgger, inverse or adversarial IL, and establishing frameworks for contact-rich tasks and human interactive tasks in unstructured environments like construction sites \cite{Argall2009SurveyLfD}. This section will focus on the evolution of imitation-based robot skill learning for construction installation tasks.  

The first IL in construction study focused on geometry-adaptive assembly. For ceiling tile installation, \citeN{liang2020teaching} captured human demonstrations (video and trajectory) and translated them into robot trajectories that generalize to varying grid layouts, illustrating the promise of IL for geometry-varying, quasi-repetitive work \cite{liang2022trajectory}. Subsequent efforts targeted collaborative and long-horizon site work with more robust IL methods. Li et al. (2023) \citeN{LiZou2023_AEI_GAIL_Construction} leveraged Generative Adversarial Imitation Learning (GAIL) to learn collaborative behaviors that better handle compounding error than pure behavior cloning in construction scenarios \cite{HoErmon2016_GAIL}. To reduce the physical burden of repeated full-scale demonstrations, \citeN{duan2025training} built an intuitive VR teleoperation pipeline that collects high-quality demonstrations via real-time hand-pose control; their training stack pretrains with behavior cloning and fine-tunes with GAIL and Proximal Policy Optimization  (PPO), improving data efficiency under limited demonstration budgets. \citeN{liuexploiting} also developed a VR-based demonstration system for long-horizon construction tasks such as bricklaying. With the proposed constant-skew motion planning, the robot can learn repetitive construction tasks with a single demonstration.

To further reduce demonstration collection workload, \citeN{yu2024cloud} proposed a cloud-based HIL architecture, where workers provide immersive virtual demonstrations that are decomposed into reusable sub-skills, stored in a federated “skill cloud.” Keyframe-centric HIL has also been proven effective. \citeN{li2024teleoperation} introduced a teleoperation-driven keyframe-based hierarchical IL framework that learns at sparse, semantically meaningful waypoints, improving trajectory controllability and generalization for repetitive construction tasks. Beyond the policy layer, IL also improves with the data collection methods, especially methods integrating Building Information Modeling and digital twins. \citeN{wang2023automatic} presented an intuitive high-level motion sequencing method “taught” from demonstrations collected from a BIM-robot loop, which maximized the usage of developments in the construction information workflow.

\subsection{Hierarchical Reinforcement Learning}
As outlined in the Introduction, several challenges remain before robots can reliably take over long-horizon material manipulation. While some barriers are tied to on-site practicalities, many can be addressed through advances in mechanism design and system-level learning. Recent progress in temporal and behavioral abstraction includes the options framework \cite{sutton1999options,precup2000temporal}, latent skill models for unsupervised or offline skill discovery \cite{eysenbach2019diayn}, and diffusion-based policies for decision-making and visuomotor control \cite{janner2022planningDiffusion,chi2023diffusionpolicy}. The most critical bottleneck, however, is sample-efficient learning that captures human manipulation skills from limited demonstrations and sparse on-robot experience. This section reviews recent advances in sample-efficient methods that ground our proposed approach. 
Hierarchical models have gained traction because they promote skill reuse and provide clear temporal abstraction \cite{barto2003hrlreview}. For instance, ROMAN performs hierarchical behavior cloning with a mixture-of-experts gating architecture and demonstrates robust long-horizon manipulation from demonstrations \cite{triantafyllidis2023roman}. Likewise, CRISP is a hierarchical RL algorithm that induces curricula from a handful of expert demonstrations to reproduce long-horizon, trajectory-level skills in simulation and transfer them to real tasks \cite{singh2023crisp}. Despite these successes, most systems omit multimodal inputs, e.g., combining end-effector trajectories with tactile/force signals. Accordingly, one of our research goals is to test whether augmenting demonstrations with force data and richer modalities measurably improves HRL training and performance on long-horizon and complex installation tasks.

\subsection{Vision-Language-Action Model}
For assembly and installation tasks, VLA models have been progressing for better performance \cite{zhong2025survey,shao2025large}. As an instance, Perceiver-Actor (PerAct) and VisuoMotor Attention (VIMA) showed that 3D voxelization and multimodal prompts increase data efficiency and long-horizon task success, crucial for assembly-like precision \cite{shridhar2023perceiver,jiang2022vima}. The combination of the VLA and IL approach has also shown promising results. RoboFlamingo demonstrated simple VLM adaptation via imitation learning for cost-effective deployment \cite{li2025roboflamingo}. Diffusion-policy backbones further stabilize precise, multi-modal action generation under latency and noise \cite{chi2023diffusionpolicy}.

The high-performance benefits from both model architecture and the evolution of training techniques. Early “language-for-robotics” systems are generally in the form of Vision-Language Models (VLM). They couple an LLM planner with grounded skills. For example, SayCan/PaLM-SayCan used an LLM for high-level sequencing while value functions gated feasible robot skills \cite{ahn2022can}. This decoupled reasoning from low-level control but required curated skill libraries, which largely increased the programming and training efforts.  

To tackle this problem, end-to-end skill learning was proposed. For example, the Robotics Transformer (RT-1) line then scaled end-to-end visuomotor policies. RT-1 introduced a language-conditioned, image-to-action transformer trained on 130k real episodes (700+ tasks) collected with a 13-robot fleet; a TokenLearner compresses visual tokens and a decoder emits discretized actions \cite{rt22023arxiv}. RT-2 co-fine-tunes a VLM with robot data and casts actions as text tokens, letting web knowledge to be transferred into action generation \cite{rt22023arxiv}. As a result, the above evolution of the model architecture serves to improve data efficiency and zero-shot generalizability of robot learning. For example, scaling real data (RT-1) lifted success and zero-shot generalization on long-tail household skills through language-conditioned policies \cite{rt22023arxiv}.   

In addition to the model itself, the training and fine-tuning techniques were also improved in the past year to reduce the re-programming and domain adaptation efforts. The first technique is to co-fine-tune VLMs with robot data. RT-2 trains on web V+L tasks and robot trajectories, sharing parameters so semantics inform action generation; actions are tokenized as text \cite{rt22023arxiv}. RoboFlamingo lightly fine-tunes an off-the-shelf VLM with imitation learning, showing strong data-efficiency \cite{li2025roboflamingo}. 

The second training innovation is the cross-embodiment adaptation. RT-1-X/RT-2-X align action spaces to a canonical 7-DoF end-effector representation and discretize actions, enabling one policy across many robots; this supports quick per-robot adaptation without bespoke redesign (O’Neill et al. 2024). Octo explicitly advertises rapid finetuning to new observation/action spaces \cite{Octo2024}. Prompting and instruction-tuning were also very helpful. VIMA shows that prompt design (text + visual goals/constraints) serves as a powerful conditioning channel, improving generalization to novel compositions—relevant for variable installation sequences \cite{jiang2022vima}. SayCan remains influential for skill selection with value-guided constraints in long-horizon tasks \cite{ahn2022can}. 

Overall, VLA models have attracted considerable attention. Yet, their potential for construction tasks remains unexplored. This paper aims to provide a systematic evaluation of these emerging models and benchmark their performance against widely used imitation learning (IL) and reinforcement learning (RL) methods in construction robot programming. Task success rate will serve as the primary evaluation metric. In addition, given that the construction industry often suffers from limited data resources \cite{yu2024cloud}, the authors further compare the sample efficiency of VLA, IL, and RL approaches to better assess their practicality and applicability in construction robotics.

\section{Methodology}
This work addresses three major challenges in construction robot learning: (i) the need for efficient demonstration collection tools, (ii) the complementary trade-offs between precision-oriented and generalization-oriented learning models, and (iii) the requirement for safe and realistic execution in long-horizon material installation tasks. To tackle these issues, we contribute: (1) two lightweight data-collection interfaces: a mouse-driven joint teleoperation system for HRL demonstrations and a keyboard-based interface for VLA data collection; (2) an adhesion-task environment pattern: a MuJoCo simulation environment providing a reusable template for material installation; and (3) two complementary models: an HRL framework that emphasizes action-level detail and a VLA framework that emphasizes instruction-following and few-shot generalization. The details of the models are discussed in the following subsections. 

\subsection{Demonstration Collection System}
\begin{figure}[H]
  \centering
  \includegraphics[width=0.8\linewidth]{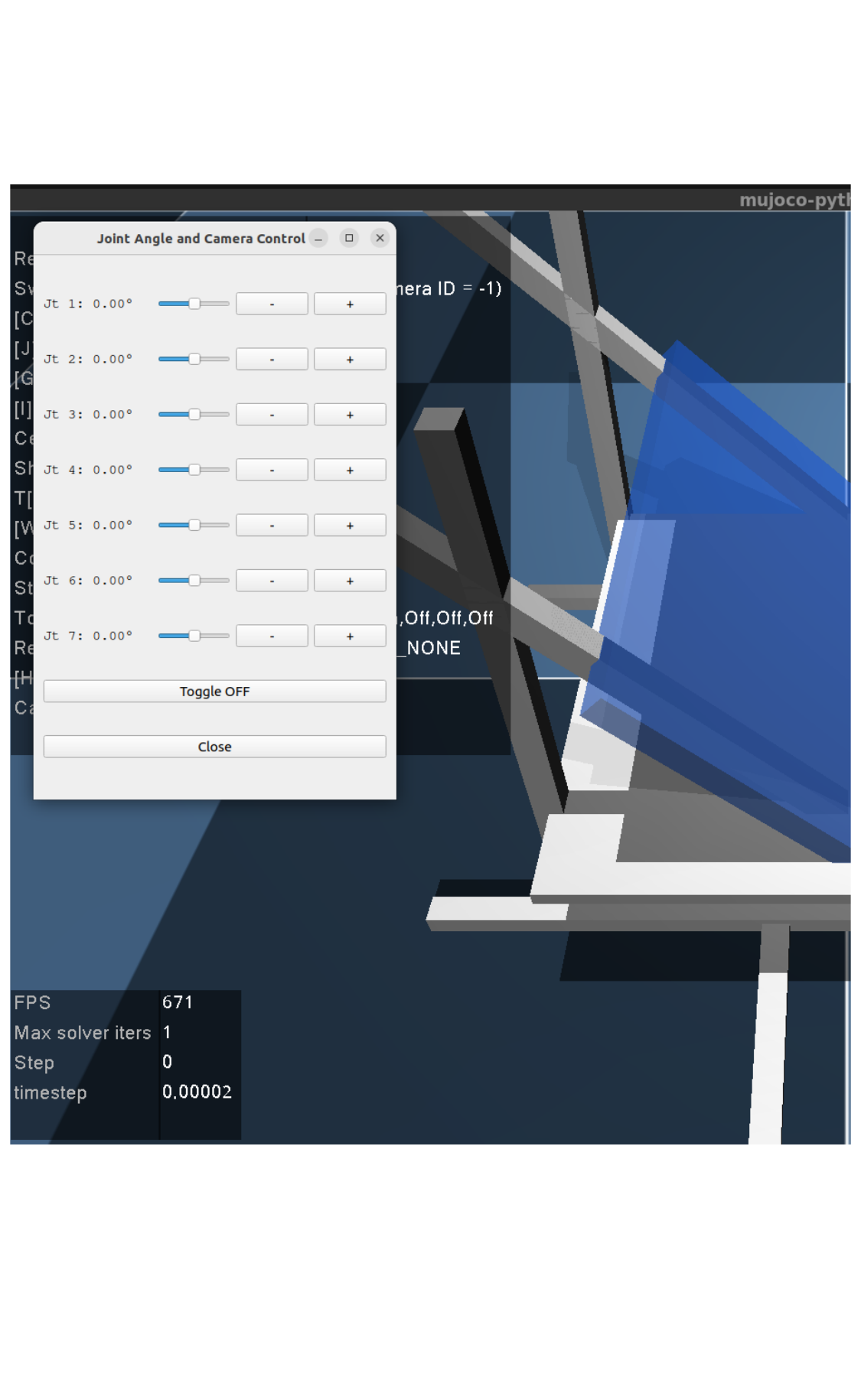}
  \caption{Joint Control Sliders in the MuJoCo Environment.}
  \label{fig:mujoco_joint}
\end{figure}

\begin{figure}[H]
  \centering
  \includegraphics[width=\linewidth]{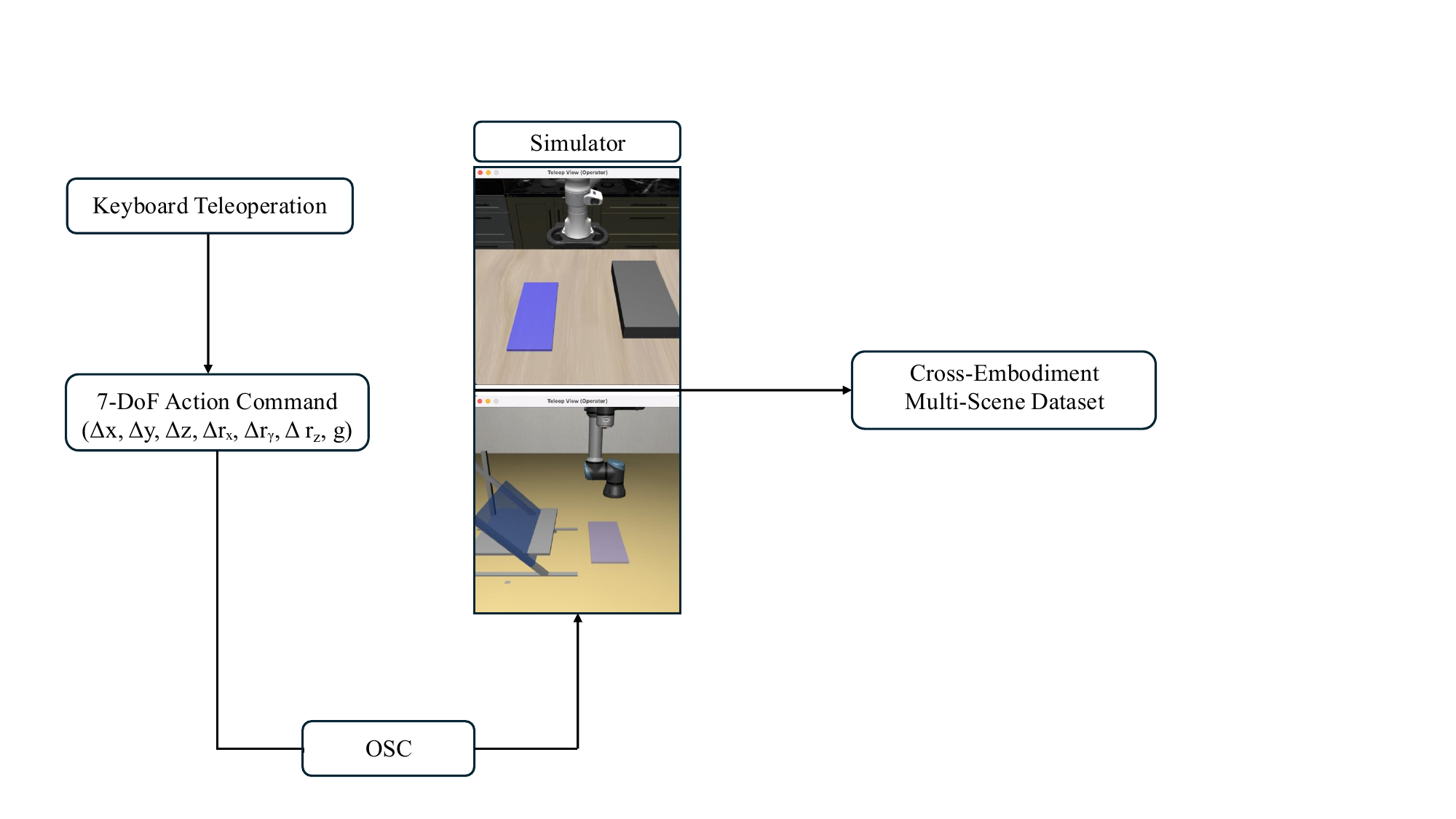}
  \caption{VLA data collection procedure.}
  \label{fig:vla_data_collector}
\end{figure}

We develop two different controller interfaces for teleoperating the robot in the simulator: (i) joint angle control with sliders and (ii) end-effector control with keyboard. Figure \ref{fig:mujoco_joint} and Figure \ref{fig:vla_data_collector} show the simulation environment along with the slider interface. Both teleoperating methods have been proven effective for robot installation skill learning tasks \cite{liang2022trajectory,chandramouliconstruction}.

For the joint control interface, users can move the robot's joints with a precision of 0.1 rad to teleoperate the robot and achieve the desired motion. This interface will return three variables: the 7-DoF pose of the manipulated object, the joint states of the robot, and the collision force imposed on the robot. The sampling frequency of the environment can go as high as 100 Hz. However, when used in learning, a filtering window of 25 will be applied twice to remove unnecessary noise from a high sampling frequency. 

Operators use the joint-control interface by manually moving each joint to produce the desired motion as shown in Figure \ref{fig:mujoco_joint}. In this study, the demonstrators first practiced pick-up and place motions to become familiar with the interface. Formal data collection began only after the demonstrator reached a satisfactory level of proficiency and could minimize collision forces during installation. To increase variety and ensure a normal distribution in the training data, the demonstrator’s operating method was not constrained. Each data collection episode lasted approximately three minutes. The duration varies based on the demonstrator and the installation methods adopted. 

For VLA-oriented demonstrations, we employ a keyboard teleoperation interface that drives a 7-DoF end-effector action at 20~Hz. Operators controlled the end-effector in Cartesian space using key presses. This type of data collection does not require any user training and is very convenient and suitable for large-scale data collection. Figure \ref{fig:vla_data_collector} shows the data collection procedure and the simulation environment. Each control step specifies a continuous vector $[\Delta x,\Delta y,\Delta z,\Delta r_x,\Delta r_y,\Delta r_z,g]$, where translation and rotation deltas are issued via dedicated keys, and the gripper/adhesion command $g$ is triggered by a separate key. After the action vector is obtained, the authors employ an Operational Space Controller (OSC) to convert these Cartesian deltas into low-level joint torques/commands for the robot. This ensures that the end-effector follows the desired 6-DoF motion while simultaneously executing the gripper action, providing smooth and physically consistent demonstrations. 
Episodes are recorded in MuJoCo with synchronized agent-view RGB frames (256$\times$256), wrist-camera images, proprioceptive states, and 7-DoF actions; each trajectory is paired with a natural language instruction and stored in an RLDS-compatible format to support downstream fine-tuning \cite{ramos2021rlds}.

\subsection{Learning Models}
\subsubsection{Hierarchical Reinforcement Learning}
\begin{figure}[H]
  \centering
  \includegraphics[width=\linewidth]{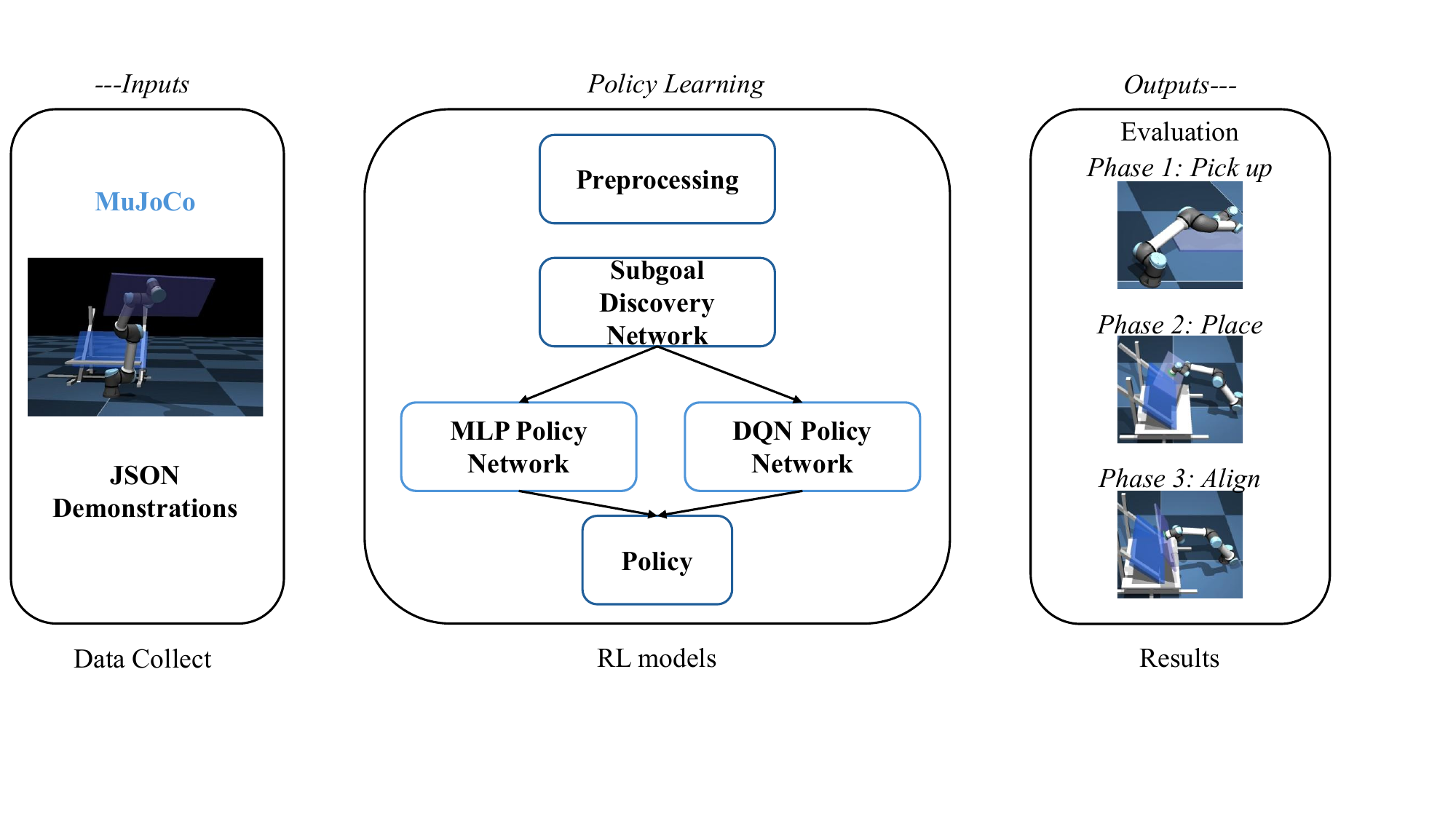}
  \caption{Informational architecture of HRL models.}
  \label{fig:DQN_meth}
\end{figure}

We adopt two RL baselines to contextualize results against the vision–language policy: a compact Multi-Layer Perceptron (MLP) used as a direct policy and a Deep Q-Network (DQN)–based imitation alternative. Both consume proprioceptive and task-relevant state and are trained on the same dataset to enable a fair comparison with the VLA setting. The information flow for the RL part is shown in Figure \ref{fig:DQN_meth}.

The MLP policy maps a low-dimensional observation to joint-space actions. The MLP policy is a three-layer fully connected network with ReLU between linear layers, taking force, joint state, and object state as observation and outputting joint actions. The DQN imitation model uses the shaped reward for the pick-up phase and explores a suitable reward for the installation phase from the demonstration data:
\begin{equation}
R_1 = -1*\alpha* D_1
\end{equation}

where $D_1$ measures the instantaneous distance to the target, $\alpha>0$ is a trained parameter scales the penalty.
The DQNs generally penalize distance to the goal, and is trained with a learning rate of $5\times 10^{-4}$ after comparing computational performance. The observation at time $t$ is
\begin{equation}
\mathbf{o}_t \;=\; \big[\,\mathbf{f}_t,\, \mathbf{q}_t,\, \mathbf{s}_t\,\big]
\end{equation}
where
$\mathbf{f}_t$ denotes force and torque readings, $\mathbf{q}_t$ the robot joint state, and $\mathbf{s}_t$ the object state. Actions are produced by a three-layer fully connected network with ReLU activations between linear layers,
\begin{equation}
\mathbf{u}_t \;=\; \pi_{\phi}(\mathbf{o}_t)
\end{equation}
which learns joint-space commands directly from $(\mathbf{o}_t,\mathbf{u}_t)$ supervision. The shallow depth and ReLU nonlinearity provide a simple inductive bias while mitigating vanishing gradients and offering mild regularization through sparse activations.

\begin{figure}[H]
  \centering
  \includegraphics[width=\linewidth]{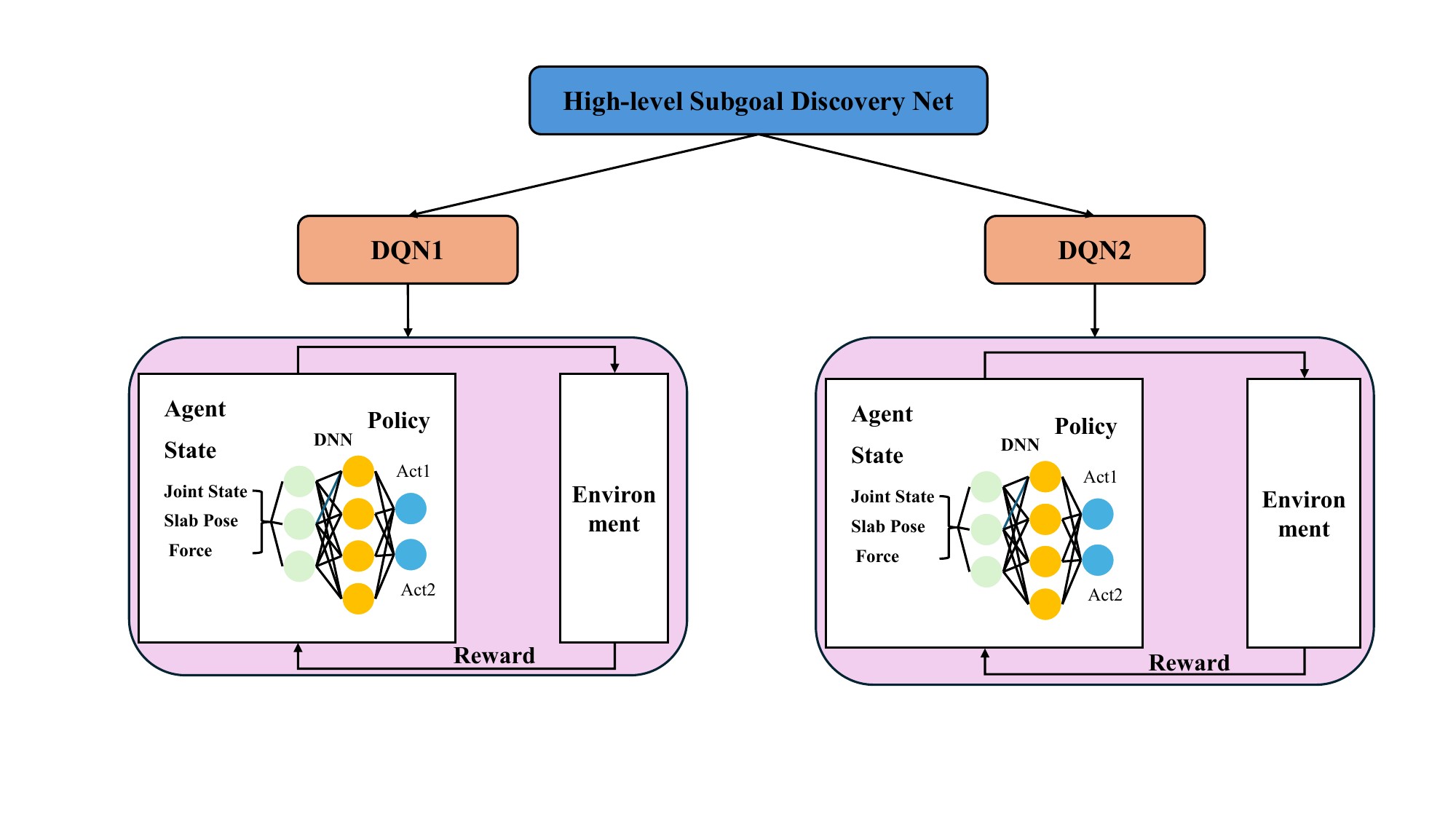}
  \caption{Architecture of RL models.}
  \label{fig:dqn_arch}
\end{figure}

As shown in Figure \ref{fig:dqn_arch}, the DQN-based imitation model estimates state–action values $Q_{\psi}(\mathbf{o},\mathbf{u})$ and selects actions by $\arg\max_{\mathbf{u}} Q_{\psi}(\mathbf{o},\mathbf{u})$. 

 Training follows a standard temporal-difference loss:
\begin{equation}
\mathcal{L}_{\mathrm{DQN}}(\psi)
\;=\;
\mathbb{E}\!\left[
\Big(
R_t
+ \gamma \max_{\,\mathbf{u}'} Q_{\bar{\psi}}(\mathbf{o}_{t+1},\, \mathbf{u}')
- Q_{\psi}(\mathbf{o}_t,\, \mathbf{u}_t)
\Big)^{2}
\right]
\end{equation}
with a learning rate of $5\times 10^{-4}$ selected after comparing computational performance across candidate settings. Together, these two HRL baselines offer complementary perspectives: direct policy regression via the MLP and value-based selection via the DQN—against which the VLA approach is evaluated.

\subsubsection{Vision-Language-Action Model}
\begin{figure}[H]
  \centering
  \includegraphics[width=\linewidth]{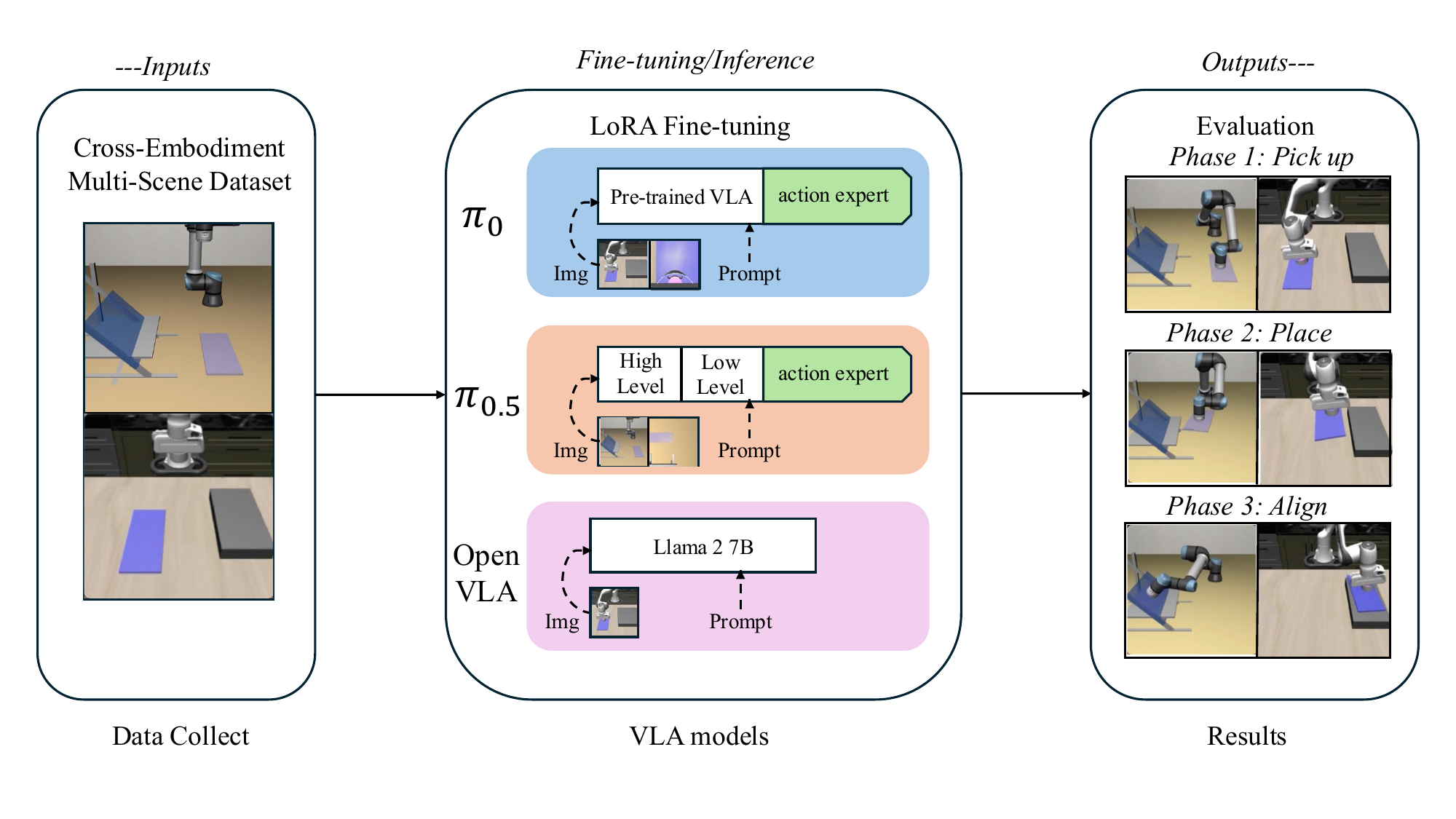}
  \caption{Informational architecture of Vision-Language-Action (VLA) models.}
  \label{fig:vla_arch}
\end{figure}

For the VLA, we benchmark three different models: OpenVLA \cite{kim24openvla}, $\pi_0$ \cite{black2024pi_0}, and $\pi_{0.5}$ \cite{intelligence2025pi_}. Figure \ref{fig:vla_arch} illustrates the procedure of the VLA robot system. First, we collect demonstration data and create a cross-embodiment multi-scene dataset (Figure \ref{fig:vla_data_collector}). Instead of a large amount of demonstration data, we simply select two different scenes and collect a few demonstration data to achieve sample efficiency. Then, three different VLA models are fine-tuned based on the demonstration data. Finally, the model is evaluated in the testing scenes.

\begin{figure}[H]
  \centering
  \includegraphics[width=\linewidth]{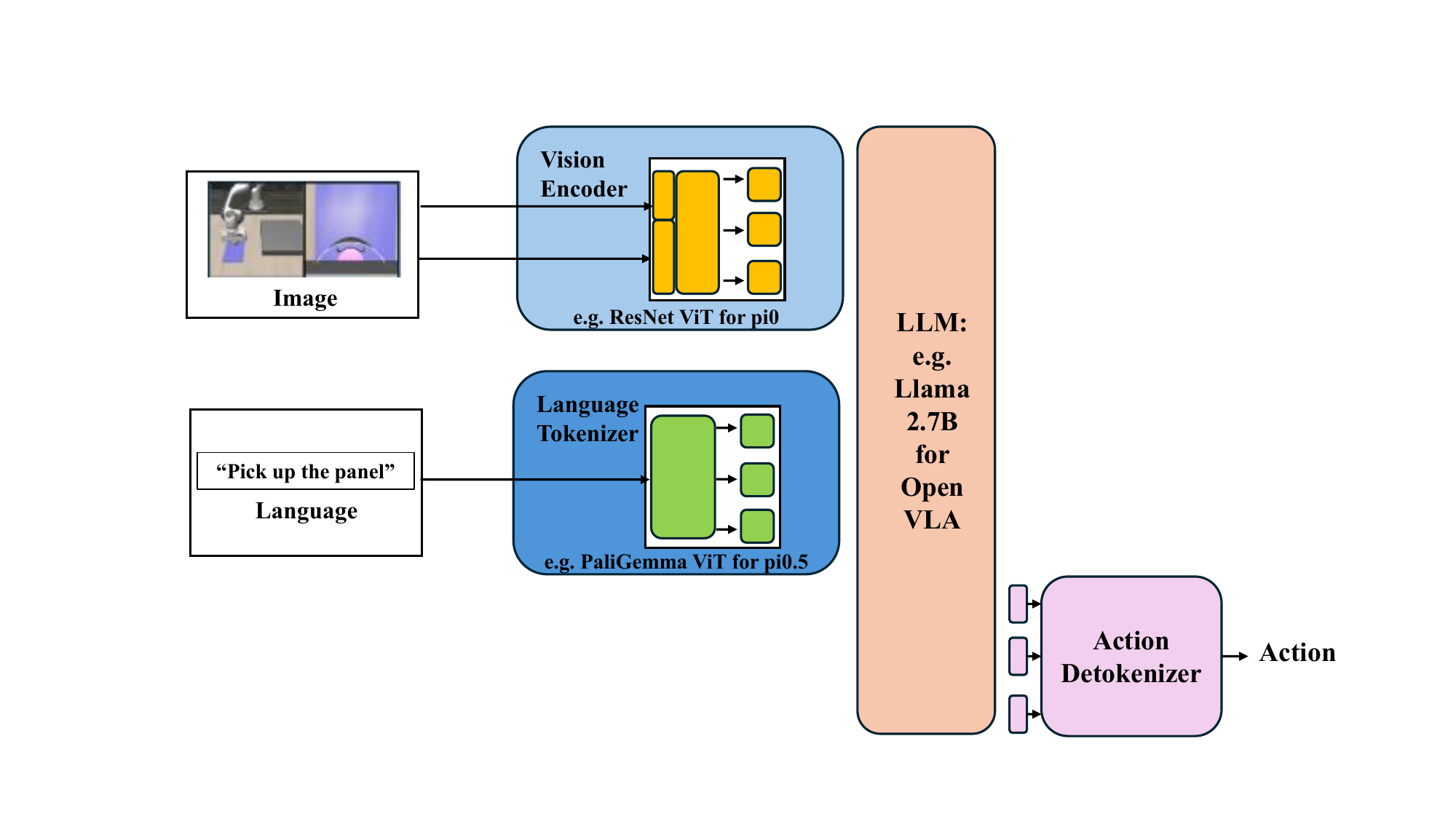}
  \caption{Detailed architecture of Vision-Language-Action (VLA) models.}
  \label{fig:vla_structure}
\end{figure}

In the VLA model fine-tuning and inference process, we use an end-to-end vision–language policy with a frozen backbone and a lightweight control head. As can be seen in Figure \ref{fig:vla_structure}, the backbone encodes the most recent RGB frames together with a task prompt (and, when available, proprioception and the previous action) into a fused representation:
\begin{equation}
\mathbf{f}_t \;=\; \Phi_{\mathrm{VLA}}\!\left(\mathcal{I}_{t-k:t},\, \mathcal{P}_t,\, [\,\mathbf{o}_t,\, \mathbf{a}_{t-1}\,]\right) \in \mathbb{R}^{d}.
\end{equation}
The control head attends to $\mathbf{f}_t$ and outputs continuous actions in end-effector space. Given $\mathbf{f}_t$, the policy predicts a 7-DoF or 6-DoF end-effector delta and a gripper or adhesion command:
\begin{equation}
\mathbf{a}_t \;=\; \big[\,\Delta \boldsymbol{\xi}_t,\; s_t\,\big] \;=\; \mathcal{H}_{\theta}(\mathbf{f}_t),
\qquad
\Delta \boldsymbol{\xi}_t \in \mathbb{R}^{7},\; s_t \in \mathbb{R}.
\end{equation}
where $\Delta \boldsymbol{\xi}_t$ encodes Cartesian position and orientation deltas, $s_t$ is squashed to the actuator range. For precise, low-latency control, $\mathcal{H}_\theta$ is a conditional flow model trained with a flow-matching objective. 

Let $\mathbf{x}_1\equiv\mathbf{a}_t$ be the demonstration action and $\mathbf{x}0\sim p_0$ a simple base. With an interpolant $\psi_t(\mathbf{x}0,\mathbf{x}1)$, the head parameterizes a time-indexed vector field $v\theta(\cdot)$ and minimizes
\begin{equation}
\mathcal{L}_{\mathrm{FM}}(\theta)
\;=\;
\mathbb{E}\Big[
  \big\|
    v_{\theta}\!\big(\psi_t(\mathbf{x}_0,\mathbf{x}_1),\, t \,\big|\, \mathbf{f}_t\big)
    - \partial_t \psi_t(\mathbf{x}_0,\mathbf{x}_1)
  \big\|_2^2
\Big].
\end{equation}
In inference, actions are produced in a single pass (or by integrating a very shallow ODE), which keeps latency low. An autoregressive head is possible but is not used here due to higher decoding delay and exposure bias. The control head is modular and decoupled from the backbone. During adaptation, only the head (and optional low-rank adapters) is trained to attend to the fused features, which enables rapid transfer to new tasks without changing the backbone.

The three representative VLA models that we used for the long-horizon construction task---$\pi_{0}$, $\pi_{0.5}$, and OpenVLA---expose complementary capability boundaries in their architectural differences. This lets us probe the performance of different architectures in construction adhesion scenarios. First, $\pi_{0}$ uses a PaLI-Gemma-3B vision-language backbone, which couples a ViT image encoder with a Gemma-2B decoder. Time-aligned RGB images from an agent view and a wrist camera, together with the text prompt, are embedded and projected into the language embedding space, then processed by the VLM. An action expert module was adopted after the backbone, proprioception, and the previous action are routed directly to the action expert module, which models a short horizon of continuous actions via flow matching: the policy predicts a 50-step action chunk of end-effector deltas and gripper commands, enabling high-frequency control up to roughly 50~Hz on dexterous contact tasks. The mixture-of-experts design routes image-text tokens to the VLM weights and state-action tokens to the action expert, which is on the order of 300M parameters; combined with the 3B VLM, $\pi_{0}$ is about 3.3B parameters in our setup. By reducing control latency, this design facilitates finer-grained manipulation, rendering it especially well-suited to our construction tasks. We initialize from the open-source released pretrained $\pi_{0}$-base checkpoint and adapt it to our construction task with parameter-efficient LoRA fine-tuning, which inserts lightweight adapters into both the VLM backbone and the action expert while keeping most weights frozen.

Second, $\pi_{0.5}$ preserves the same I/O and control interface as $\pi_0$ but is co-trained on heterogeneous data to strengthen open-world generalization and long-horizon stability. Thus, we are able to benefit from the stronger pretrained $\pi_{0.5}$-base checkpoint. At inference, the model first produces a high-level subtask in natural language with the autoregressive pathway, then realizes it with a continuous 50-step action chunk via the same flow-matching action expert. This ``reason-then-act'' decoding retains $\pi_{0}$'s high-rate continuous control while improving robustness across new homes and object distributions, which is the reason for selecting $\pi_{0.5}$ for longer sequences and more complex assembly tasks. We adopt the same fine-tuning procedure for $\pi_{0.5}$ as for $\pi_{0}$.

Finally, OpenVLA serves as a contrastive baseline that removes both multi-view RGB and the continuous expert head. It builds on a Prismatic-7B VLM: features from DINOv2 and SigLIP are concatenated, projected to the Llama-2-7B token space, and the policy predicts actions as discrete tokens by discretizing each action dimension into 256 bins with next-token training. In our configuration, we follow the official pre-training guideline to feed a single agent-view image, so OpenVLA tests a single-view, discrete-action regime against $\pi_{0}$/$\pi_{0.5}$'s multi-view, chunked continuous control. The released checkpoints report around 7.5B parameters, including the visual encoders. We initialize from the pretrained OpenVLA checkpoint and adapt it to our domain using the same LoRA fine-tuning strategy.







\section{Evaluation}
The evaluation was conducted in three stages. The first stage compares MLP and DQN-based imitation policy to select the stronger baseline. The second stage compares three VLA models to find the most robust VLA model. The third stage benchmarks the selected RL baseline against VLA.

\medskip
\noindent\textbf{Stage I: DQN vs.\ MLP (preliminary).}
The comparison considers: (i) model performance, (ii) generalization, and (iii) a pickup experiment. The better model from this stage is used as the HRL baseline for the next stage.

\medskip
\noindent\textbf{Stage II: $\pi_0$ vs.\ $\pi_{0.5}$ vs.\ OpenVLA (preliminary).}
The comparison considers: (i) model performance, (ii) generalization, and (iii) a long-horizon pick and place experiment.

\medskip
\noindent\textbf{Stage III: VLA vs.\ DQN (benchmark).}
This stage adopts two types of metrics. First, computational and sample-efficiency measures are used to understand the effort required to achieve satisfactory performance. Second, a robot experiment is conducted on a dual-phase hierarchical panel installation task, evaluating the pickup phase to assess performance in a straightforward movement phase and a dexterity-required installation phase.

\subsection{Experiment Setup}
For the robotics component of our study, we integrate the policies trained in the HRL stage with the MuJoCo physics simulator. Specifically, both the MLP and DQN agents export their learned policies in the form of model checkpoints. These policies are then loaded into MuJoCo, where they serve as controllers for the robot’s motion. The exported models output discrete action signals, which are translated into joint torques and position commands within the simulator to direct the robot's motions.

For VLA, we adapt three models, $\pi_{0}$, $\pi_{0.5}$, and OpenVLA, to the long-horizon construction task. All models are initialized from their publicly released pretrained checkpoints. All experiments are conducted in the MuJoCo physics simulator, where we follow settings from LIBERO \cite{liu2023libero} and Robosuite \cite{robosuite2020} to construct two workplaces: a ground-level scene equipped with a UR5e arm and a desk-level scene equipped with a Franka Panda arm, as shown in Figure \ref{fig:vla_data_collector}. All robots are equipped with an adhesion gripper.
Within these environments, we collect the Cross-Embodiment Multi-Scene dataset and fine-tune all models on this dataset using parameter-efficient LoRA adapters. The dataset consists of 100 demonstration data for the desk scenario and 200 demonstration data for the ground scenario. Performance is measured under the standard success-rate protocol across repeated rollouts (see Stage II in the Results section).

\subsection{Results}
\subsubsection{Stage I: DQN vs.\ MLP}
\begin{figure}[H]
  \centering
  \includegraphics[width=\linewidth]{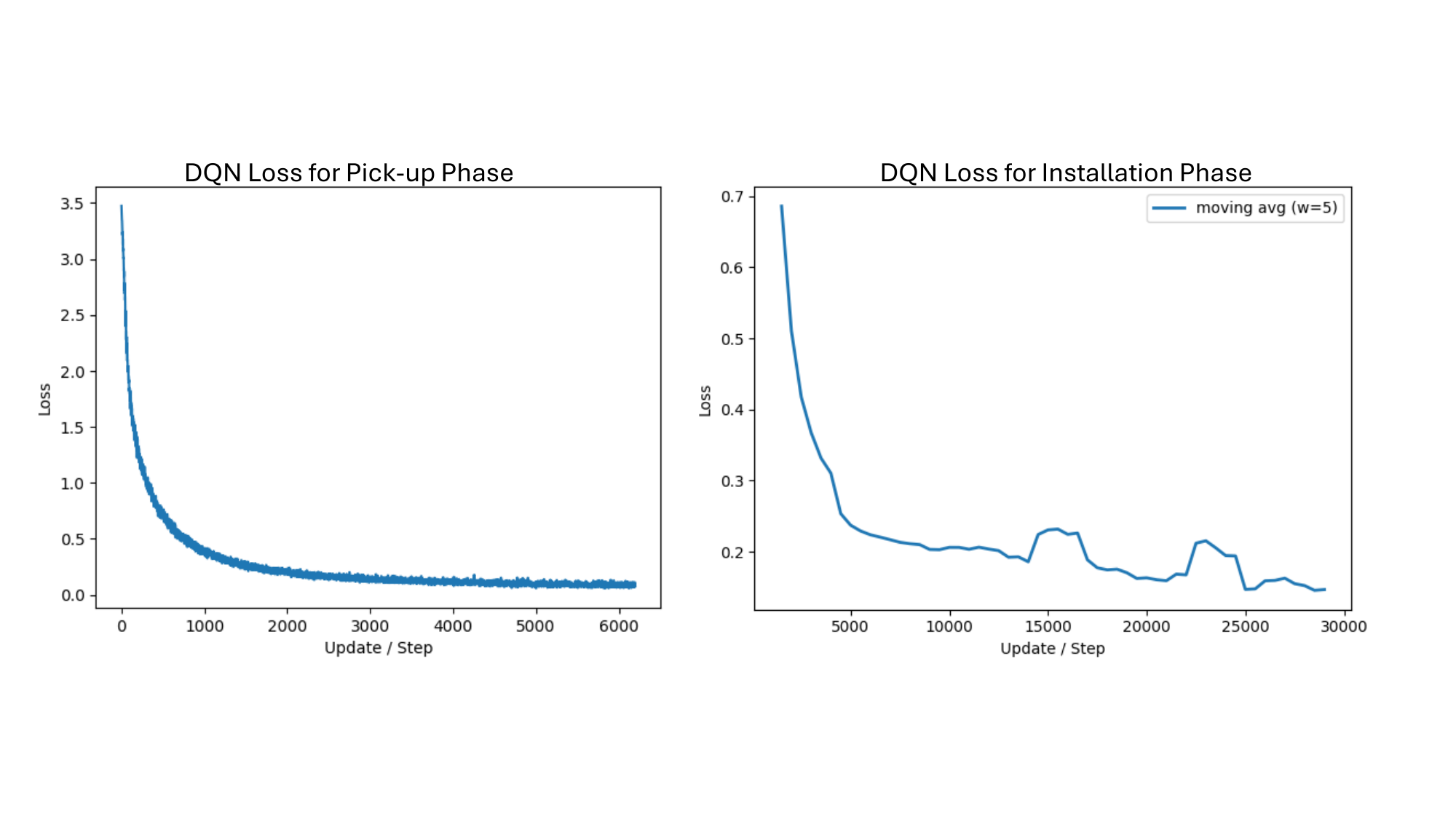}
  \caption{Training loss curves of DQN reward models.}
  \label{fig:HRLloss}
\end{figure}

The designed or explored reward functions successfully imitate the human's behavioral skills. As shown in Figure \ref{fig:HRLloss}, both networks will have minimal loss at the end of the training. The Stage I preliminary evaluation of MLP and DQN for robot task execution is also provided in Table~\ref{table:mlp_dqn_results}. Given the same 100 demonstrations, DQN achieves a pickup success rate comparable to MLP. The MLP, however, displays clear overfitting. In terms of sample efficiency, MLP fits the fastest. It is likely due to its simpler architecture and smaller parameter count. This suggests it needs less data to model the training distribution but generalizes less reliably. Given the similar success rates but greater robustness, we select DQN for subsequent experiments.

\subsubsection{Stage II: $\pi_0$ vs.\ $\pi_{0.5}$ vs.\ OpenVLA}
\begin{figure}[H]
  \centering
  \includegraphics[width=\linewidth]{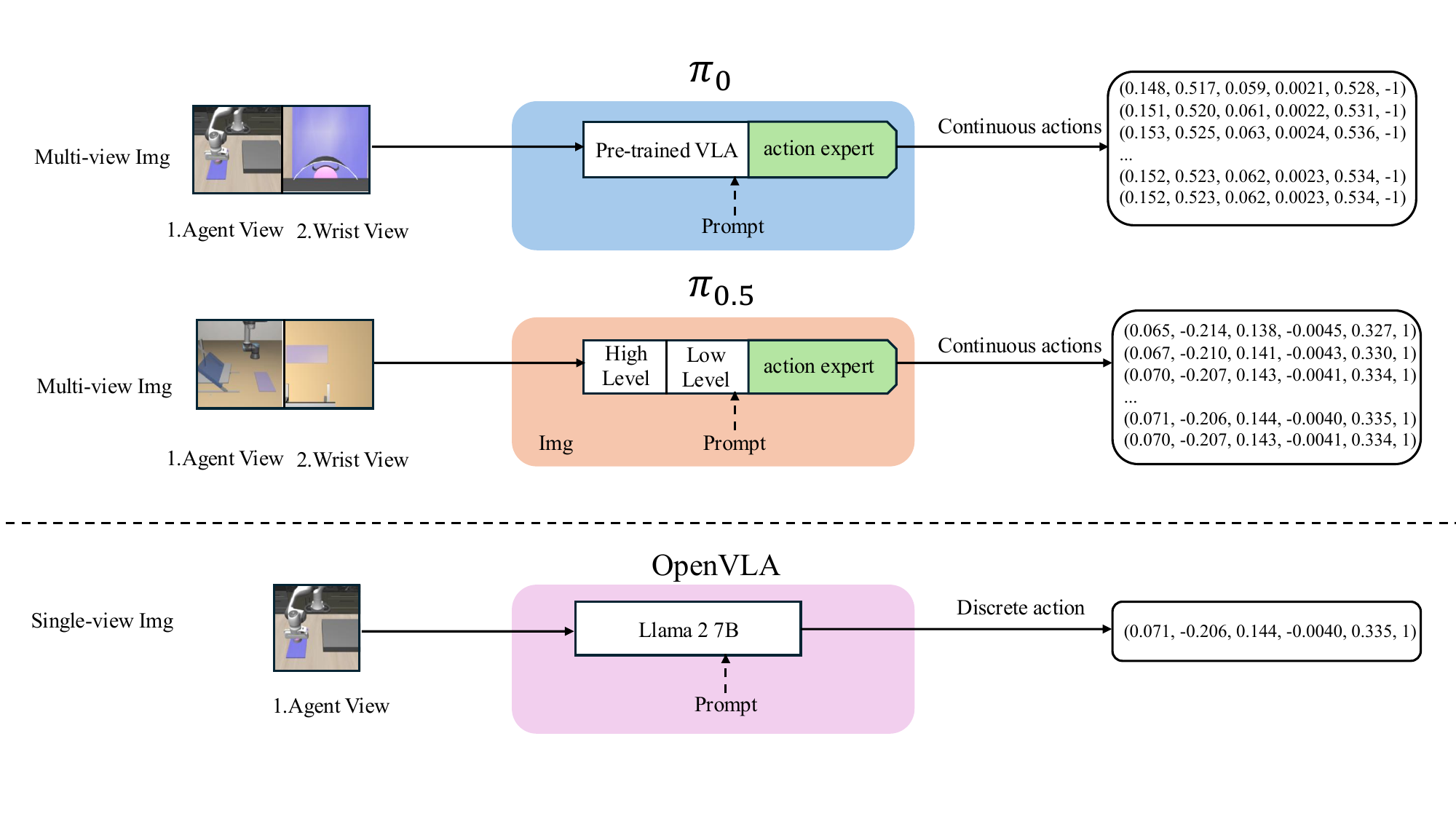}
  \caption{Comparison of $\pi_{0}$, $\pi_{0.5}$, and OpenVLA architectures and action outputs.}
  \label{fig:diff}
\end{figure}

For the Stage II preliminary evaluation of VLA, we follow LIBERO's evaluation protocol: each model is tested over 50 rollouts in environments that match their training settings to obtain success rates across repeated trials. During each testing period, the position of the panel is subjected to slight perturbations, in contrast to the fixed placement used in DQN evaluation. This setup is designed to assess the generalization capability of the VLA model. Table~\ref{table:vla_results} shows the performance of the VLA models. In the desk workplace, where the task requires picking up a panel from a desk and placing it onto a stand, $\pi_{0}$ achieves solid success rates, benefiting from multi-view perception and a continuous action head that enables smooth control. The ground workplace presents a more challenging scenario, requiring the panel to be lifted from the ground and inserted into a tilted frame; under this setting, $\pi_{0.5}$ achieves consistently high success, as it retains the advantages of $\pi_{0}$ while providing more powerful pre-trained models trained with new methods. By contrast, OpenVLA fails in both tasks due to two key limitations: (i) it only has access to the agent-view camera, which makes it difficult to perceive thin, flat panel and to align precisely for grasping without wrist-camera observations; and (ii) it lacks an expert continuous action head, instead outputting short discrete action tokens, which prevents smooth trajectory generation. These limitations cause unstable pickup attempts and poor alignment, ultimately leading to zero success across all stages. Figure~\ref{fig:diff} shows the difference between two different architectures in inference. 

\begin{figure}[H]
  \centering
  \includegraphics[width=1.1\linewidth]{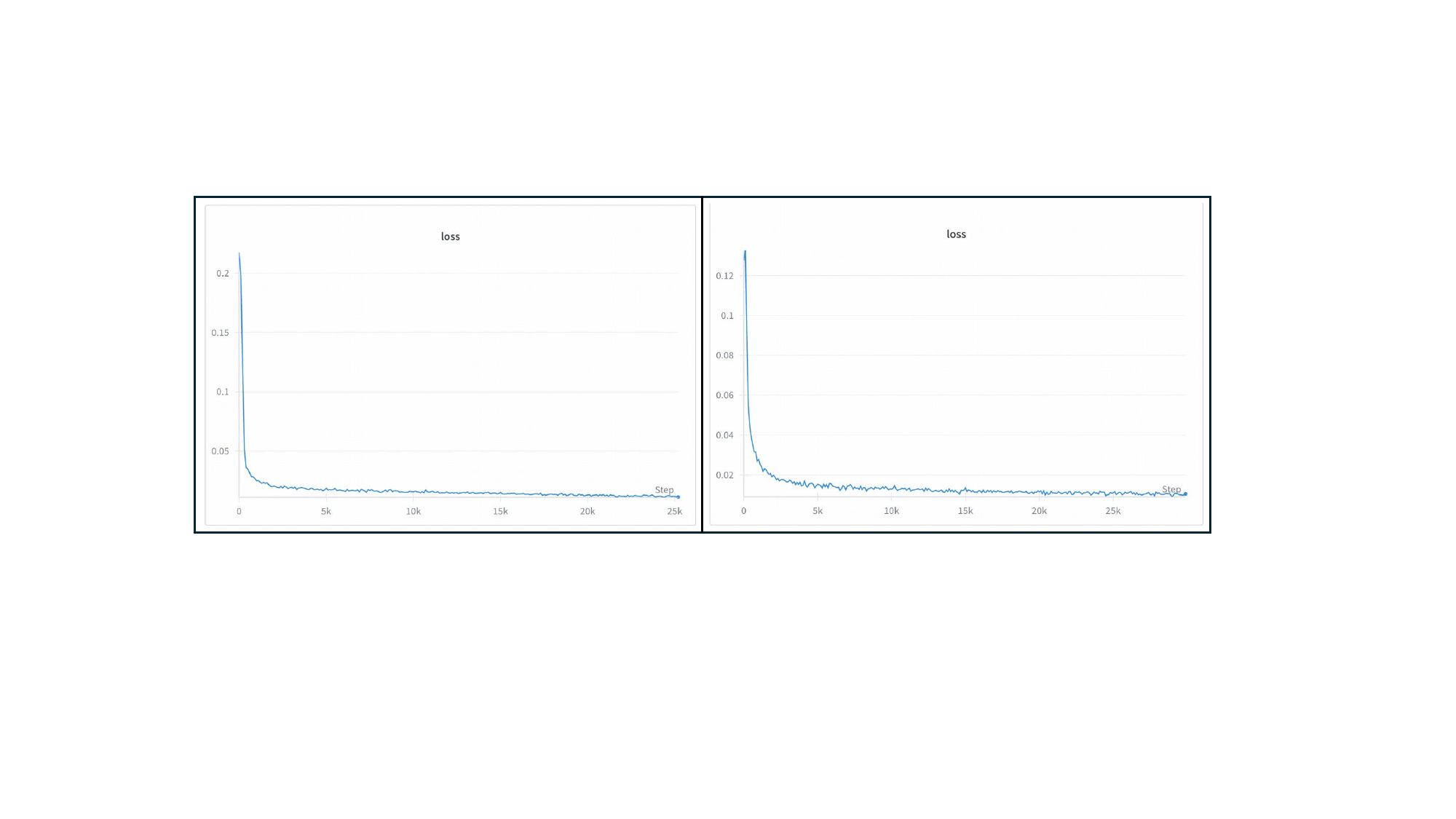}
  \caption{Training loss curves of $\pi_{0}$(left) and $\pi_{0.5}$(right) with AdamW.}
  \label{fig:loss_curves}
\end{figure}

\begin{figure}[H]
  \centering
  \includegraphics[width=\linewidth]{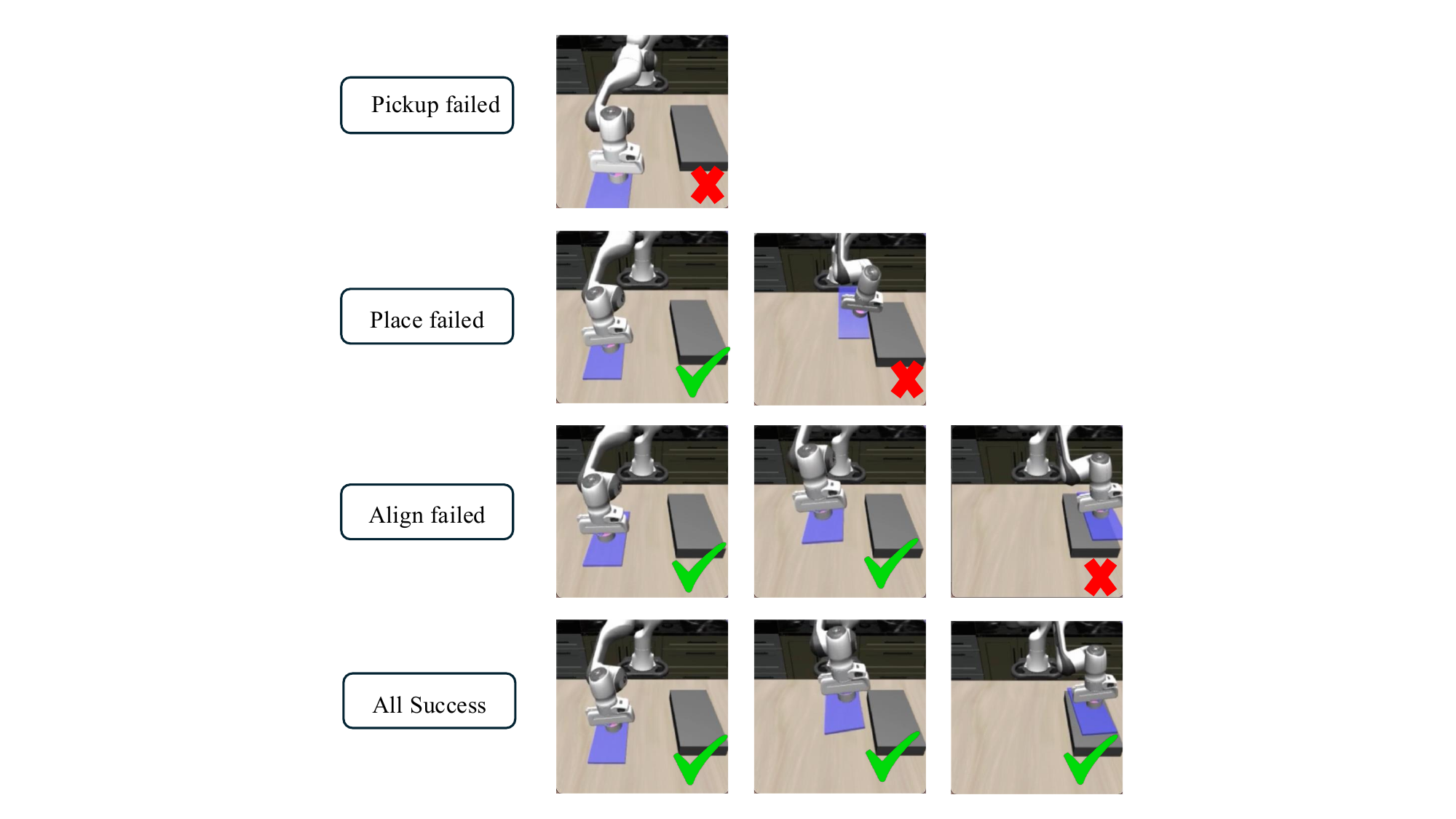}
  \caption{$\pi_{0}$ performance on desk workspace.}
  \label{fig:pi0_result}
\end{figure}

\begin{figure}[H]
  \centering
  \includegraphics[width=\linewidth]{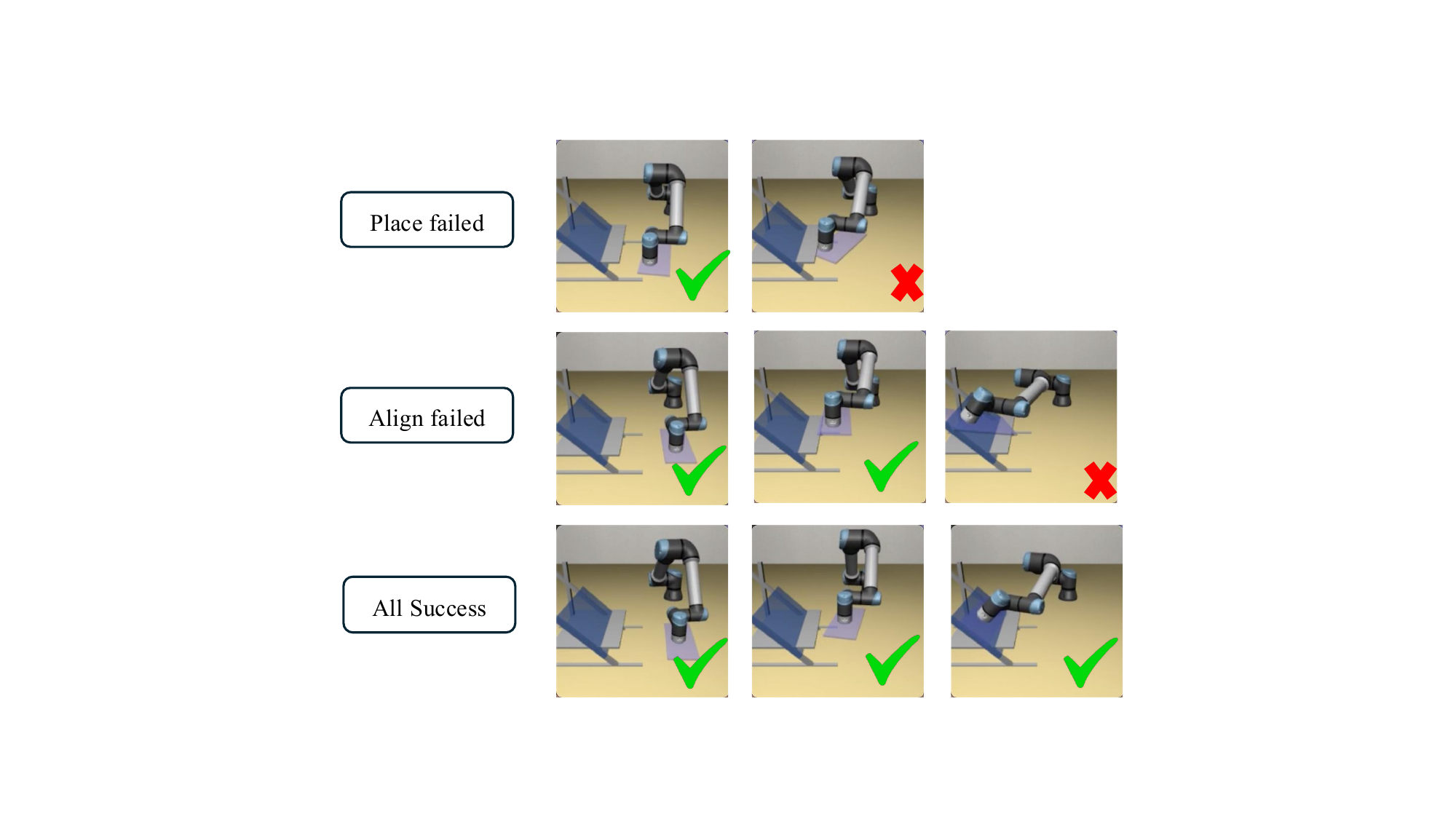}
  \caption{$\pi_{0.5}$ performance on ground workspace.}
  \label{fig:pi0.5_result}
\end{figure}

\begin{figure}[H]
  \centering
  \includegraphics[width=\linewidth]{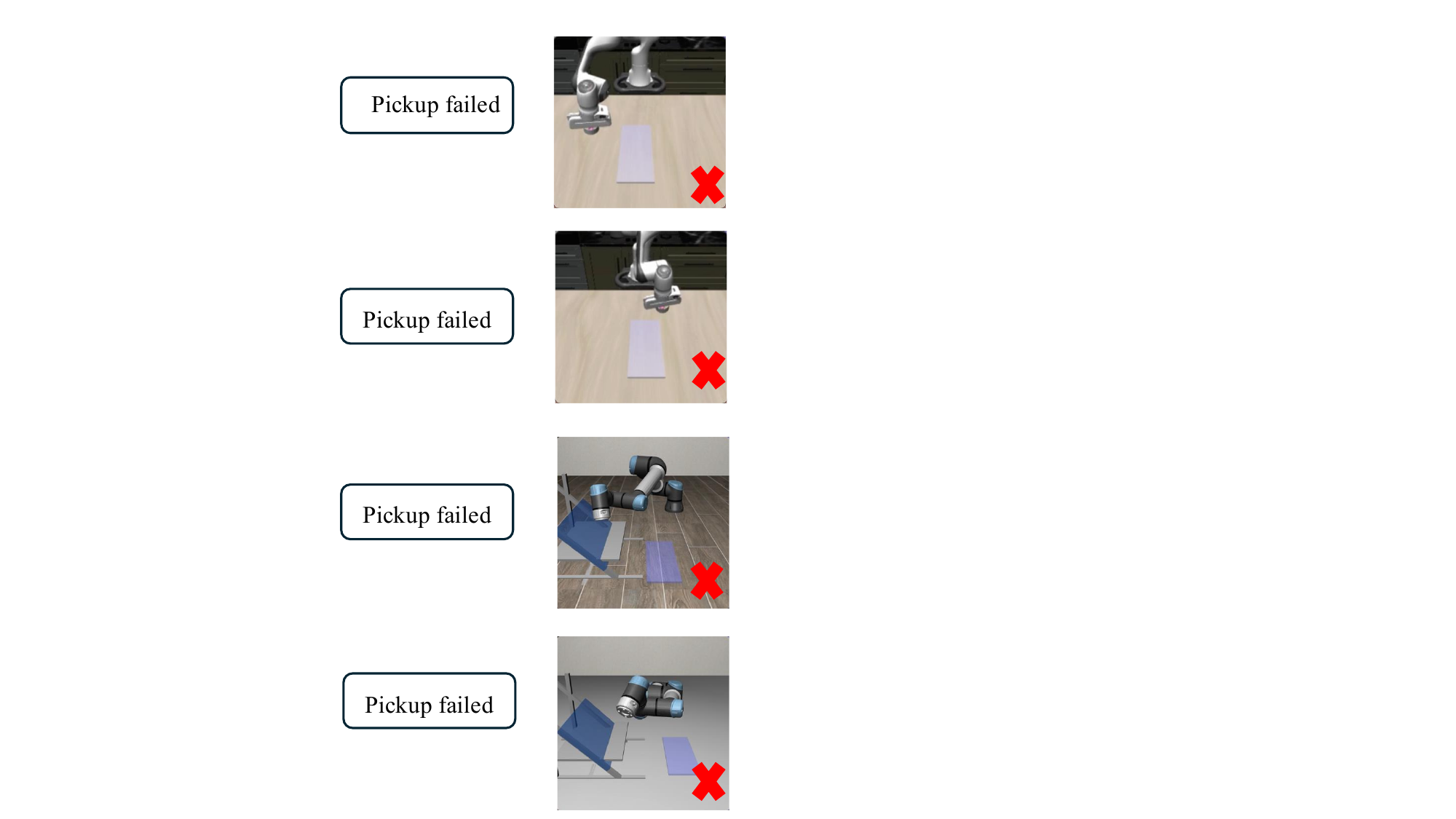}
  \caption{OpenVLA performance on desk and ground workspace.}
  \label{fig:openvla_result}
\end{figure}

All models were optimized using the AdamW optimizer. Figure~\ref{fig:loss_curves} shows the training loss of $\pi_{0}$ and $\pi_{0.5}$. Both exhibit rapid early decreases followed by stable convergence, with $\pi_{0.5}$ achieving a slightly lower asymptotic floor (around $0.01$). For OpenVLA, although the full loss curve was not logged due to training configuration differences, partial records suggest convergence to a similar range near $0.01$, indicating stable optimization despite its failure to succeed in rollout evaluations. The performance results in terms of success and failure cases of the three models are presented in Figure~\ref{fig:pi0_result}, Figure~\ref{fig:pi0.5_result}, and Figure~\ref{fig:openvla_result}. The robot manipulation sequence of two success cases on desk and ground workspaces can be found in Figure~\ref{fig:seq_desk} and Figure~\ref{fig:seq_ground}. The experiment videos are provided in the supplement materials.

\begin{table}[h]
\caption{Results of MLP and DQN models performance}
\label{table:mlp_dqn_results}
\centering
\begin{tabular}{lccc}
\hline
& Pickup success rate & Average joint angle error (rad) & Overfit \\
\hline
DQN & 100\% & 0.184 & no \\
MLP & 100\% & 0.020 & yes \\
\hline
\end{tabular}
\end{table}

\begin{figure}[H]
  \centering
  \includegraphics[width=\linewidth]{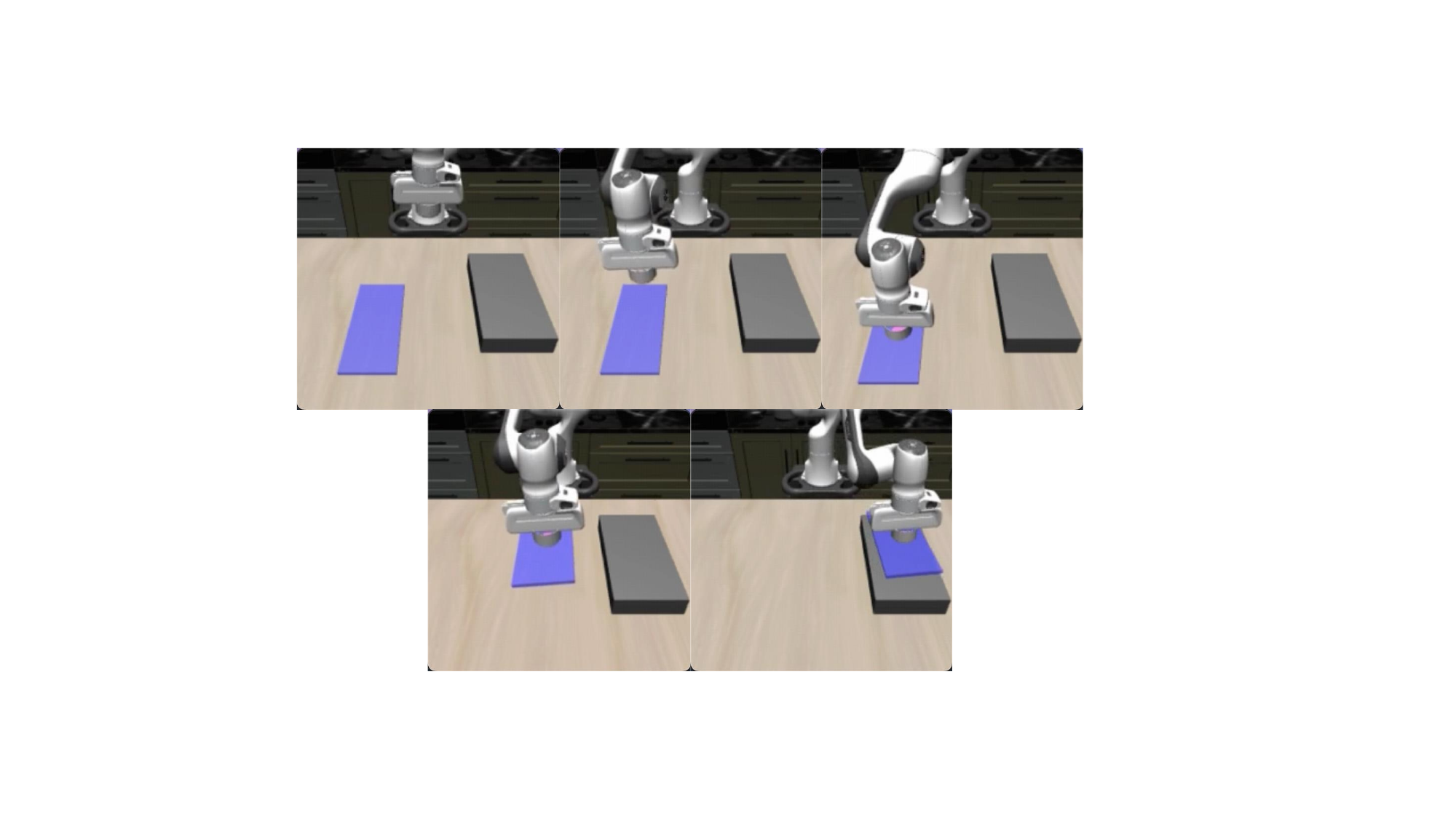}
  \caption{Robot sequence of desk workspace.}
  \label{fig:seq_desk}
\end{figure}

\begin{figure}[H]
  \centering
  \includegraphics[width=\linewidth]{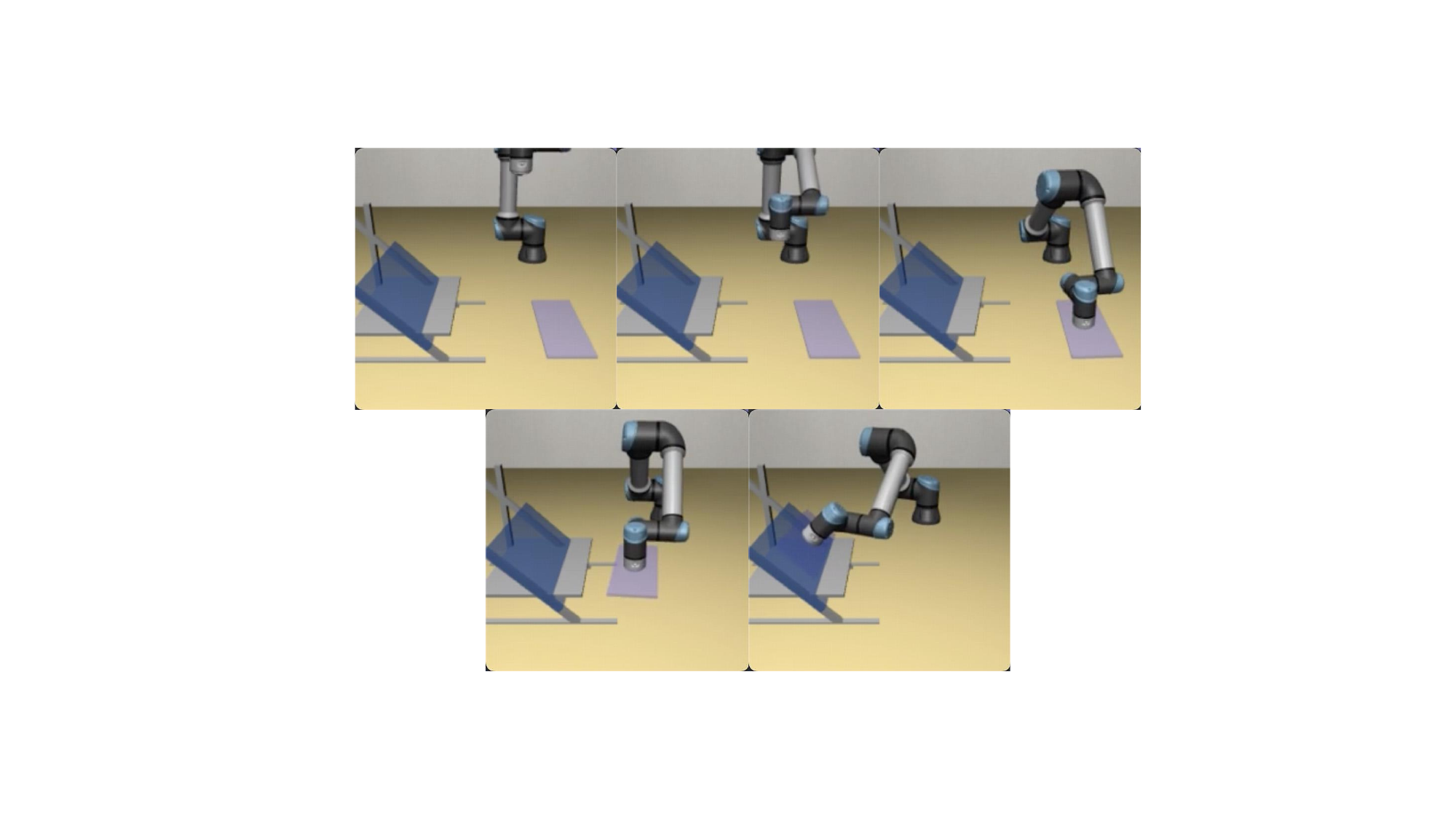}
  \caption{Robot sequence of ground workspace.}
  \label{fig:seq_ground}
\end{figure}

\begin{table}[h]
\caption{Results of VLA model performance}
\label{table:vla_results}
\centering
\begin{tabular}{lcccc}
\hline
Model & Workplace & Pickup success rate & Place success rate & Align success rate \\
\hline
$\pi_{0}$   & desk   & 60\%  & 44\% & 32\% \\
$\pi_{0.5}$ & ground & 100\% & 96\% & 80\% \\
OpenVLA     & ground/desk & 0\%   & 0\%  & 0\%  \\
\hline
\end{tabular}
\end{table}

\subsubsection{Stage III: VLA vs.\ DQN}
To compare VLA and DQN, we use the results of the pickup experiment listed in Table~\ref{table:mlp_dqn_results} and Table~\ref{table:vla_results}. Both methods achieve 100\% success rate. DQN requires less training data (100) than VLA fine-tuning (200) to achieve the same success rate. VLA also requires large-scale pre-trained models. However, VLA demonstrates more robustness and generalizability since it has some perturbations in the panel location. To summarize, this section provides the systematic evaluation results for VLA and RL on three axes: the training sample efficiency, robustness, and few-shot generalizability. RL requires less data when training from scratch, yet VLA provides the strongest few-shot performance.

\section{Discussion}
This paper systematically compares two recent advanced approaches to robot skill learning: reinforcement-based imitation learning and VLA. This paper employs hierarchical structures in both to improve sample efficiency. In VLA, hierarchy is inherent to the model. In the RL setting, a specifically designed HRL architecture is used, comprising an MLP subgoal discovery network and two Q-networks trained separately for the pick-up and installation tasks.

Based on computational studies and robot execution, the results indicate that VLA exhibits stronger generalizability and few-shot capability with minimal warm-up data; it attains a 60\% ($\pi_0$) and 100\% ($\pi_{0.5}$) success rate in the pick-up phase. By contrast, DQN requires additional noise during tuning to achieve robust behavior, which increases the overall workload.

The authors acknowledge a misalignment in parts of the experimental design. For VLA, demonstrations provide vision and action inputs, whereas for DQN, the observation space uses trajectory and force. This difference is considered both important and consequential for performance. For VLA, the most effective inputs align with its original training modality (vision and action). Although force inputs could improve VLA, implementing them would require architectural modifications \cite{yu2025forcevla}, which are beyond the present scope that focuses on baseline model differences and benchmarking VLA. For DQN, reward shaping that penalizes distance to the goal can impose limitations. When relying on trajectory-only demonstrations, the learned policy may pass above the grid and risk collisions with construction materials. Accordingly, each model is evaluated with the most suitable and effective modalities to both enhance performance and serve the benchmarking objective.

A second limitation concerns the minimal number of demonstrations required to reach the reported performance. While the number of demonstrations is a useful indicator of sample efficiency, this work does not determine the exact minimum. To account for variable demonstration quality, increments of $10$ demonstrations are used for DQN and MLP, and the precise threshold is not measured. Even so, the observed patterns are informative: VLA is highly sample-efficient for few-shot skill transfer, whereas DQN does not achieve few-shot capability under the tested conditions.

\section{Conclusions}
This paper compared two widely used approaches to robot skill learning: reinforcement-based imitation learning and a VLA model under a hierarchical design aimed at improving sample efficiency. VLA employs an inherent hierarchical structure with three different models, while the RL side uses a dedicated HRL setup with an MLP subgoal discovery network and two Q-networks trained for the pick-up and installation tasks. Evaluation includes in three stages: a preliminary comparison between an MLP policy and a DQN-based imitation model (considering model performance, generalization, and a pick-up experiment), a preliminary comparison between three VLA models ($\pi_0$, $\pi_{0.5}$, and OpenVLA), followed by a benchmark between the selected HRL baseline and VLA using computational and sample-efficiency measures and a dual-phase panel installation robot experiment (transport and installation).

With the same $100$ demonstrations, both DQN and MLP achieved comparable pick-up success, but the MLP exhibited clear overfitting (validation loss increasing beyond training loss), leading to the selection of DQN as the more robust baseline for subsequent comparisons. In the stage II and stage III experiments, with $100$ and $200$ demonstration data in two different scenarios, VLA demonstrated strong robustness, generalizability, and few-shot capability, achieving a $60\%$ and $100\%$ pick-up success rate, whereas DQN required additional noise during tuning to obtain robust behavior, increasing the overall workload.

Two limitations were acknowledged. First, there is a modality mismatch in parts of the design: VLA was evaluated with vision and action demonstrations, while DQN used trajectory and force as observations. This choice reflects the most suitable and effective modality for each model; force inputs could benefit VLA but would require architectural modifications, which are beyond the scope here. Second, the minimal number of demonstrations needed to reach the reported performance was not determined; demonstration counts for DQN and MLP were increased in steps of $10$, and exact minima were not measured. Even so, the results reveal patterns in sample efficiency: VLA is highly sample-efficient for few-shot skill transfer, while DQN does not achieve few-shot capability under the tested conditions.

Overall, the study establishes a baseline comparison between hierarchical RL methods and VLA under matched evaluation protocols, highlights VLA’s few-shot advantages, and identifies practical considerations for HRL (e.g., noise injection during tuning). Future work will address the noted modality alignment and systematically measure minimal demonstration requirements.

\section{Data Availability Statement}

 Some or all data, models, or code that support the findings of this study are available from the corresponding author upon reasonable request.

\section{Acknowledgments}
This project was carried out without external sponsorship.
\pagebreak
%
\appendix
\bibliography{ascexmpl-new}

\end{document}